\documentclass[conference]{IEEEtran}
\IEEEoverridecommandlockouts

\usepackage{cite}
\usepackage{amsmath,amssymb,amsfonts}
\usepackage{algorithmic}
\usepackage{graphicx}
\usepackage{textcomp}
\usepackage{xcolor}
\def\BibTeX{{\rm B\kern-.05em{\sc i\kern-.025em b}\kern-.08em
    T\kern-.1667em\lower.7ex\hbox{E}\kern-.125emX}}
\begin{document}

\title{Zonal Architecture Development with evolution of Artificial Intelligence\\
}

\author{\IEEEauthorblockN{1\textsuperscript{st} Sneha Sudhir Shetiya}
\IEEEauthorblockA{\textit{IEEE Senior Member} \\
\textit{Michigan, USA} \\
sneha.shetiya@ieee.org}
\and
\IEEEauthorblockN{2\textsuperscript{nd} Vikas Vyas}
\IEEEauthorblockA{\textit{IEEE Senior Member} \\
\textit{California, USA}\\
vikas.vyas@mercedes-benz.com}
\and
\IEEEauthorblockN{3\textsuperscript{th} Shreyas Renukuntla}
\IEEEauthorblockA{\textit{PhD Student} \\
\textit{Oakland University}\\
Michigan, USA \\
shreyas.renukuntla@gmail.com}
}

\maketitle

\begin{abstract}
This paper explains how traditional centralized architectures are transitioning to distributed zonal approaches to address challenges in scalability, reliability, performance, and cost-effectiveness. The role of edge computing and neural networks in enabling sophisticated sensor fusion and decision-making capabilities for autonomous vehicles is examined. Additionally, this paper discusses the impact of zonal architectures on vehicle diagnostics, power distribution, and smart power management systems. Key design considerations for implementing effective zonal architectures are presented, along with an overview of current challenges and future directions. The objective of this paper is to provide a comprehensive understanding of how zonal architectures are shaping the future of automotive technology, particularly in the context of self-driving vehicles and artificial intelligence integration.
\end{abstract}


\section{Introduction}
\begin{figure*}[htbp]
\centering
\begin{tabular}{c|c}
\hline
\multicolumn{2}{c}{\includegraphics[width=\linewidth,height=4.0in]{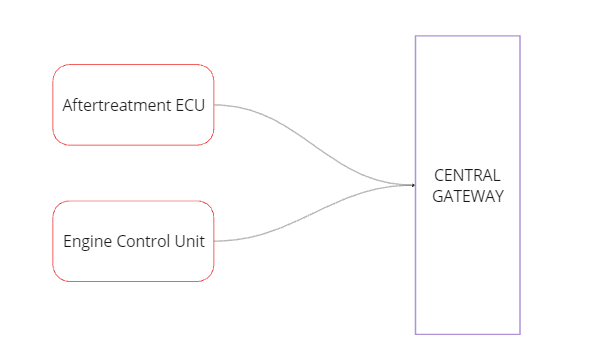}}\\
\hline
\end{tabular}
\caption{Complex vehicle architecture.}
\label{fig1}
\end{figure*}

\begin{figure*}[htbp]
\centering
\begin{tabular}{c|c}
\hline
\multicolumn{2}{c}{\includegraphics[width=\linewidth,height=4.0in]{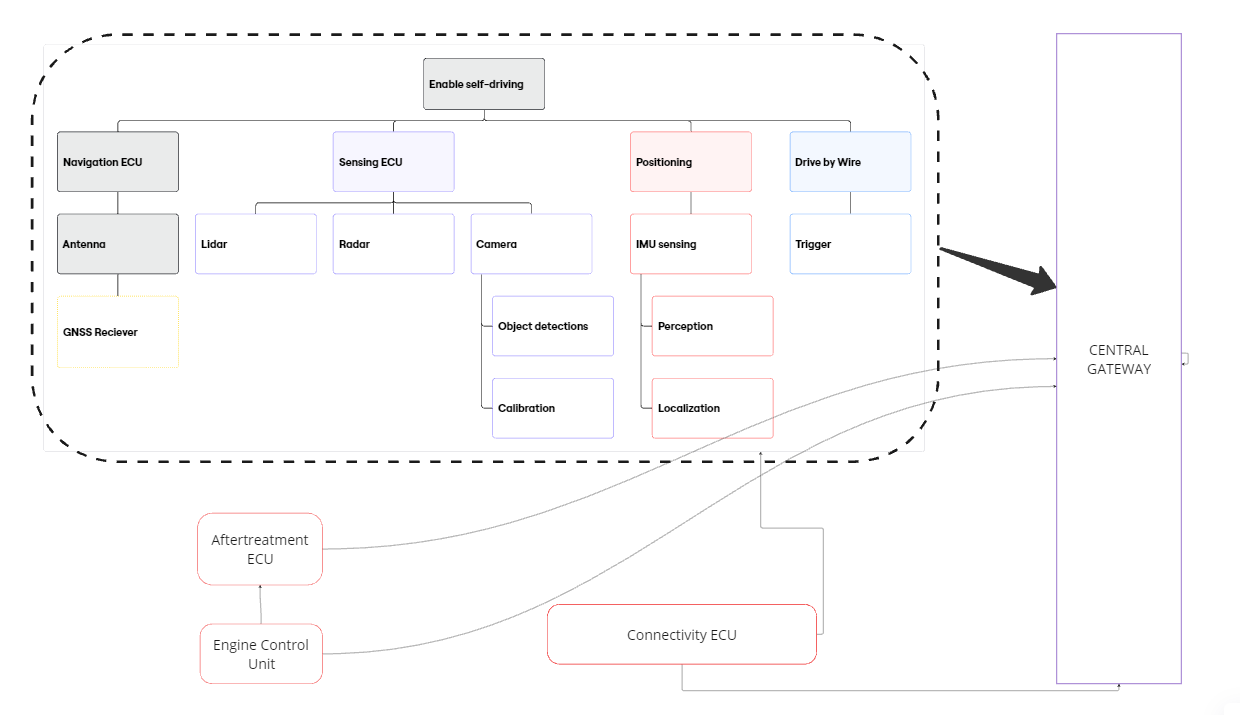}}\\
\hline
\end{tabular}
\caption{Complex vehicle architecture.}
\label{fig2}
\end{figure*}

Automotive technology has advanced exponentially over the decades. From the traditional Motor wagon\cite{b1} that was drawn before the turn of the century to today's cars equipped with self-driving technology, we see advancement both in software and hardware of the chassis. Earlier we just needed the functionality for the engine strokes and aftertreatment functions. But now we need Over The Air(OTA) Software Update\cite{b2} capabilities and various other connectivity features\cite{b3}. The complexity of the architecture is also growing parallely with this. More and more Electronic Control Units(ECUs)\cite{b4} are being embedded in the vehicle for extra features added. This is giving rise to complex connections and since all communication cant be hard wired through ethernet or CAN, there is a need to cluster the ECUs. This is where the concept of zonal architecture\cite{b5} comes into picture.
\ref{fig1} shows the internal vehicle structure with only fewer ECUs
\ref{fig2} shows the complex internal vehicle structure with many ECUs


In order to develop the next generation of Automotive vehicles, zonal architecture plays an important role. Major benefits are as follows:

\subsubsection{Performance Improvement}
When it comes to self -driving technology, perception of the environment is crucial and also the reaction times. Thus performance plays a major role.

\subsubsection{Scalability and Flexibility}
If the architecture is scalable, then the addition and elimination of sensors and actuators is feasible. Maintenance cost is huge and debugging the cause of the issue can be time consuming and also huge costs on the pocket. Thus zonal architecture helps distribute processing power across zones and we can then localize to one area.

\subsubsection{Reliability}
Redundant Chassis is a must today for every Original Equipment Manufacturer(OEM) in order to launch a product into the market. Thus fallback mechanisms if one cluster fails need to be provided which is only possible with zonal architecture.

\subsubsection{Cost effectiveness}
As highlighted above, cost saving is a major benefit of this architecture.As every ECU need not be connected to one another and need to be connected to the respective cluster which can then be decomposed within, the cost reduces significantly as wiring complexity goes down.

\subsection{Literature Survey}
The paper \cite{b6} talks about how zonal architecture where STRIDE is implemented which is a threat model to ensure security of OTA Updates in the In Vehicle Network. This paper shows the ease with which one can ensure security for software updates in complex vehicle networks with zonal architecture.\\
There is some research also done on the real -time simulation side of things where Zonal DC systems are used\cite{b7}. This paper talks about how devices in zonal architecture can be controlled using a distributed and performing data communication network. This setup with Hardware in Loop helps in better testing of the zonal architecture in consideration.\\
We can consider any requirement today in a vehicle, safety, entertainment or driver assistance; the zones where the ECUs are placed , time synchronization has to be achieved. This is achieved through CAN-FD(Controller Area Network-Flexible Datarate) and Ethernet\cite{b8}.The testing is performed on embedded boards.\\
All the three papers discussed above show the need for a Zonal Architecture in vehicles and showcase some ways how this is simulated and tested for the latest features.\\
Major Contributions of this paper are as follows:
\begin{itemize}
\item Analyse how the Automotive vehicle architecture has evolved over the years paving the way to Zonal Architectures
\item
Verification of the claim that zonal architectures are a necessity for vehicles equipped with Autonomous Driver Assistance Features.\cite{b9}
\item
Challenges in implementing Zonal architectures and effective ways to overcome the same.
\end{itemize}

\section{Neural Network necessity for driverless vehicles}\label{neural}
We already discussed how Zonal Architecture helps segregate ECUs into clusters in order to simplify complex architecture. Going a level down,let us consider one of the important components of the next generation of driverless vehicles; sensor fusion. 
Sensor fusion involves processing huge amounts of sensor data through multiple streams coming from sensors like camera, Lidar(Light Detection and Ranging), Radar(Radio Detection and Ranging), Inertial Measurement Unit(IMU), GNSS(Global Navigation and Satellite System), Antenna and ultrasonics. Neural networks are employed here to make decisions with this data to be consumed by modules of perception, sensing, planning and control. 
There are two critical tasks that Neural Networks perform by consuming the sensor data:
\begin{itemize}
    \item \textbf{Object Recognition:} On the road and in traffic, neural networks breeze through a ton of visual information. They identify objects in the process like pedestrians, vehicles, traffic lights and lane markings.
    \item \textbf{Scene Interpretation:} Once the objects are recognized, the next step is to understand the context. For this, neural networks have to learn the relationships between the various identifies objects. They can then estimate speed, predict movement patterns of vehicles and pedestrians and anticipate any possible faults and possible threats to life. This can then kick in the safety mechanisms in the vehicle to ensure proper MRMs(Minimum Risk Maneuvers).
\end{itemize}
The environment has been perceived now and the next steps are to take critical corrective actions based on the scene interpretation above.
\begin{itemize}
    \item \textbf{Steering and Navigation:} NN(Neural Networks) have the capability to determine optimal traffic flow, identify road signs and subsequently their environment. Based on this information, NN helps maintaining speed, can change lanes for smoother journey and even apply brakes to avoid collision. 
    \item \textbf{Route Planning and Optimization}: Due to their capabilities to factor in real time data, weather conditions and even designated drop of zones in higher level of Autonomy, they plan the most efficient and safest route for the vehicle.
The major advantage of these NNs is enhanced safety. With faster reaction times and more precise decision- making skills, NNs have successfully reduced accidents cause by human error by an exponential percentage. Autonomous vehicles powered by these systems can lead to safer driving experience worldwide.
\end{itemize}

\section{Zonal Architecture: A Distributed Powerhouse for Sensor Fusion}
Understanding the need to have multiple compute units in the vehicle for various driver assistance features is essential. From \ref{neural}, it is clear the necessity to have the driverless features in vehicles today. These features need complex architectures and Zonal architecture helps implement this.

Conventional car designs often use a centralized architecture, where all sensor data converges on a single powerful computer. This approach faces challenges:
\begin{itemize}
    \item \textbf{Data Overload:} The central computer can become overwhelmed by the sheer volume of data streaming from cameras, radars, lidar, and other sensors. This can lead to processing delays and hinder real-time decision-making.
\item \textbf{Wiring Complexity:} Connecting numerous sensors to a central unit requires a complex web of wires, increasing weight, maintenance costs, and the risk of malfunctions.
\item \textbf{Limited Scalability: }Adding new sensors or features becomes difficult as the central unit reaches its processing capacity.
\end{itemize}

Zonal architecture offers a smarter approach by dividing the vehicle into zones (e.g., front left, rear right). Each zone has its own dedicated processing unit responsible for:
\begin{itemize}
\item \textbf{Sensor Fusion:} The zonal controller gathers data from various sensors within its zone and performs initial processing. This reduces the data load on the central computer and enables faster fusion of sensor information.
\item \textbf{Local Decision-Making:} The zonal controller can make basic decisions based on the fused sensor data, like initiating emergency braking if an obstacle is detected within the zone.
\item \textbf{Simplified Wiring:} Sensors connect to their designated zonal controller, reducing wiring complexity and improving overall system reliability.
\end{itemize}

Advantages of using Zonal Architecture in Self-Driving Cars:\cite{b10}
\begin{itemize}
\item \textbf{Enhanced Performance:}  quicker reaction times and more accurate perception of the environment.
\item \textbf{Scalability and Flexibility}: Adding new sensors or features becomes easier as processing power is distributed across zones. This allows for customization and adaptation to different driving environments.
\item \textbf{Improved Reliability:} It offers redundancy. If one zonal controller fails, the others can still function, enhancing overall system reliability.
\item \textbf{Cost-Effectiveness:} Reduced wiring complexity and the potential for using less powerful (and cheaper) zonal controllers can lead to cost savings in vehicle production.
\end{itemize}

\subsection{Design Considerations for increased complexity}
There are certain issues which need to be highlighted which occur in any vehicle architecture upon increased complexity.\cite{b11} This can happen in a typical connected vehicle or ADAS equipped vehicle. Integration nightmare is of utmost importance. As more ECUs start to communicate with each other, its is necessary to ensure seamless data exchange and avoid possible conflicts. Using traditional CAN networks, the bus will always be busy blocking the flow of communication. Thus domain gateways become necessary for broadcasting to occur. Sometimes data synchronization issues due to incorrect PTP(point to point) sync can give rise to compatibility issues. This can eventually cause system malfunction hard to recover from.
Software management can cause huge mayhem. One example is let us assume every ECU is running its own software. This may come with its own versioning scheme as and when updated. This would cause updating and debugging software across multiple ECUs a time consuming and error-prone process.
Power management is also another issue as every additional ECU added contributes to the overall power draw. This also strains the vehicle's electrical system and reduces fuel efficiency too.
With multiple ECUs handling critical ADAS functions, ensuring overall system safety becomes paramount. A malfunction in one ECU could have cascading effects, compromising the entire ADAS system.
Design Considerations for Effective ADAS Development:
\begin{itemize}
\item \textbf{Standardization is Key:} Utilizing standardized communication protocols (like CAN FD) and software development tools simplifies integration and reduces compatibility issues between ECUs.
\item \textbf{Virtualization Power:} Leveraging virtualization technology allows multiple software applications to run on a single ECU, optimizing hardware usage and simplifying software management.
\item \textbf{Modular Design Approach:} Breaking down ADAS functionalities into smaller, modular components managed by individual ECUs improves maintainability and simplifies the development process.
\item \textbf{Prioritization and Redundancy}: Critical ADAS features should have dedicated ECUs with built-in redundancy to ensure system reliability and prevent cascading failures.
\item \textbf{Focus on Security:} As ECU complexity increases, so does the vulnerability to cyberattacks. Implementing robust security measures is crucial to protect vehicle systems from unauthorized access and manipulation.
Addressing the challenges of growing ECU complexity requires a collaborative effort. Here's what the future holds:
\item \textbf{Collaboration Between OEMs and Chipmakers}: Collaboration between car manufacturers (OEMs) and chipmakers can lead to the development of more powerful and efficient ECUs specifically designed for ADAS applications.\cite{b12}
\item \textbf{Standardization Efforts}: Industry-wide adoption of standardized communication protocols and software development tools will further streamline ECU integration and simplify ADAS development.
\end{itemize}

\subsection{Zonal Architecture for Electric Vehicle(EV) Design}
Zonal architecture is a decentralization of electric controllers to several modular zones, or hardware gateways, located at points throughout the vehicle. Devices of various functions attach to the closest gateway, rather than with its domain grouping.
With the converging automotive trends of vehicle electrification and Advanced Driver Assistance Systems (ADAS), the role of traditional automotive interconnects is evolving to meet more demanding performance requirements for key automotive application areas, including high-power interconnects specific to electric vehicles, and high-speed data interconnects required for ADAS control components which will ultimately evolve to fully autonomous vehicle systems.

The next generation of high-power automotive interconnects include those for vehicle batteries/drive trains, and other internal systems; as well as external connectors for EV charging stations designed to handle the larger vehicle batteries with higher amperages that deliver faster charging speeds. 
Advanced interconnects and wiring harnesses will be required to handle significantly increased bandwidth, multiple sensor inputs from cameras, LiDAR, and radar devices, all at higher data transfer speeds with more complex network architectures needed for ADAS and autonomous vehicle systems. 
Increased bandwidth and data transmission speeds are required for the next generation of advanced driver assistance systems, including vehicle safety devices such as sensors, cameras, radar/LiDAR, and telematics, as well as infotainment and other non-safety systems. The coaxial cables and interconnect wiring harnesses for these multiple sensors must have increased data capacity and transmission speeds to support the volume of this sensor data, especially as the automotive industry transitions from driver assisted systems to fully autonomous vehicles.

One standard automotive interconnect system involves the FAKRA (Fachkreis Automobil) connector family, developed to meet the stringent mechanical and environmental requirements of the automotive industry. These connectors are available in multiple configurations and feature keyed with color-coded designs for easy identification and to prevent mis-matching, with primary and secondary locking systems for secure connections. Rated for data speeds up to 8Gps, these connectors are ideal for developing wiring harnesses for high-speed one-way data transfer from sensors located within the vehicle, including cameras, GPS navigation, infotainment, RF keyless entry, and driver comfort systems.
For safety-critical connections that require higher speed data transfer rates, wiring harnesses using the mini-FAKRA interconnect series supports up to 28Gbs data transmission. In addition to the higher data speeds, the mini-FAKRA connectors are more compact than standard FAKRA connectors, saving space and weight in vehicle electronics systems. Wiring harnesses using these connectors are ideal for autonomous driving and critical driver assistance systems, as they can deliver high data volumes from multiple sensors, cameras, and navigation sources.
Traditional automotive electrical/electronic systems use a domain-based architecture, with specific modules, devices, and sensors wired to central ECUs (electronic control units). This approach can only scale up to a certain point before reaching limitations based on size, weight, and data transfer speeds. 
Newer automotive systems feature a zonal architecture using Ethernet cable specifically designed for automotive applications. Replacing the individual wires and nodes is an Ethernet backbone that connects all the vehicle’s systems with a software-controlled network that is capable of aggregating and transmitting data from different nodes within the vehicle, using a common protocol to support different data speeds to deliver signals to different computing locations onboard the vehicle. This zonal approach uses Ethernet switches, or gateways, at each system node to manage the increased data streams required for advanced vehicle systems, and enables local computing for less time-sensitive processing while routing time-critical safety/control systems to be channeled to the central high-speed vehicle computers. 
With the increasing number of electric vehicles and vehicles with advanced driver assist systems, the need for high-speed, high-density power and data transmission is quickly outpacing the traditional systems for internal vehicle wiring and external EV charging.
\subsection{Role of Edge Computing}
Edge computing perfectly complements zonal architectures by bringing computation closer to the data source (the sensors):
\begin{itemize}
\item \textbf{Real-Time Processing:} Edge computing empowers zonal controllers to process sensor data in real-time, enabling faster decision-making for ADAS features like collision avoidance or lane departure warnings.
\item \textbf{Data Filtering and Pre-Processing:} Edge devices can perform initial data filtering and pre-processing, reducing the amount of data that needs to be sent to a central computer for further analysis. This optimizes bandwidth usage and reduces processing load on the central unit.
\item \textbf{Improved Security:} Sensitive data processing can be done at the edge, minimizing the need for data transmission across the vehicle network, potentially reducing security risks.
Edge computing and zonal architectures are poised to become a powerful duo in the future of automotive design. As technology continues to develop, we can expect:
\item \textbf{Advancements in Edge Hardware:} More powerful and efficient edge devices specifically designed for automotive applications will emerge.
\item \textbf{Standardization of Edge Computing Platforms:} Standardized platforms will simplify the development and integration of edge applications within zonal architectures.
\item \textbf{Collaboration Between Industries: }Collaboration between automotive manufacturers, chipmakers, and software developers will foster the creation of innovative edge computing solutions for next-generation vehicles.
\end{itemize}

\section{Impact on vehicle diagnostics and procedures}
Troubleshooting and diagnosing faults is an important aspect when it comes to hardware in the vehicle. For Classical ECUs , there are set standards and methodologies. As complex ECUs are introduced the faults may occur across interconnected ECUs and will not be localized. This will take additional cost and time as well for technicians to debug the cause, design specialized tool for each domain and potentially will increase repair times.

Diagnostic functionalities are majorly employed when flashing the ECU.Fault codes are limited for fixed errors.  When the error count is unique to each ECU, additional software based mechanisms are needed to diagnose the ECU. In recent years, there has been evolution of Adaptive Autosar stack \cite{b13} and Service Oriented Vehicle Diagnostics \cite{b14} standard that enable diagnostics for complex architecture of ECUs.
Zonal architecture helps in this by providing simplified diagnostics through a consolidated view of sensor data and system health within the zone. Faults can be identified quickly this way thus streamlining the diagnostic process.
Standard communication protocol can be used across zones to simplify software update and maintenance. This enables single tool for all diagnostic activities simplifying the job of a technician.
Fault can be isolated to specific zones. This reduces debug time minimizing component replacements.
Major benefits are reduced downtime of the vehicle due to faster diagnostics and quicker repairs. Improved Repair efficiency, enhanced system transparency as system health can be viewed easily, design of easy zonal specific diagnostic tools and focus on OTA updates without in person visits to manufacturing unit.
\section{Implications on vehicle power distribution and management}
Conventional car architectures rely on a centralized power distribution system. The battery supplies power to a central junction box, which then distributes it to various (ECUs) throughout the car. This approach has limitations:
\begin{itemize}
\item \textbf{Heavy Cabling:} Long runs of thick cables are needed to connect the central junction box to all ECUs, adding weight and complexity to the vehicle.
\item \textbf{Limited Scalability:} Adding new features or ECUs often necessitates additional wiring, increasing weight and potentially exceeding the capacity of the central distribution system.
\item \textbf{Single Point of Failure:} A malfunction in the central junction box can disrupt power supply to critical systems.

Zonal architecture decentralizes power distribution. The vehicle is divided into zones (e.g., front, rear, cockpit) with dedicated power distribution modules (PDMs) \cite{b15} closer to the ECUs they serve. This offers several advantages:
\item \textbf{Reduced Wiring Complexity: }Shorter cable runs within zones decrease weight and simplify wiring.
\item \textbf{Improved Scalability:} Adding new features or ECUs within a zone requires less complex wiring modifications compared to the centralized approach.
\item \textbf{Enhanced Reliability:} A failure in one zonal PDM affects only that zone, minimizing the risk of a complete power outage.
\end{itemize}

\subsection{Smart Power Management with Zonal Architecture:}
Zonal architectures pave the way for smarter power distribution and management strategies:
\begin{itemize}
\item \textbf{Smart Fuses: }Traditional fuses can be replaced with smart fuses or Power MOSFETs (Metal-Oxide-Semiconductor Field-Effect Transistors) that offer more granular control over power delivery and can react faster in case of overloads, protecting components.
\item \textbf{Voltage Regulation:} Zonal PDMs can regulate voltage levels within each zone, optimizing power efficiency for different components.
\item \textbf{Battery Management Integration}: Zonal architecture allows for better integration with battery management systems, enabling features like regenerative braking and improved battery health monitoring.
\end{itemize}

\subsection{Challenges of Zonal Architecture for Power Distribution}
\begin{itemize}
\item \textbf{Heat Management: }Distributing power across multiple PDMs can create thermal management concerns. Efficient heat dissipation strategies need to be implemented.
\item \textbf{	Standardization}: Standardization of connectors and communication protocols between zonal PDMs and ECUs is crucial for easier integration and maintenance across different car models.
\end{itemize}

Despite these challenges, the future of power distribution in vehicles seems to be headed towards zonal architectures. As technology advances, we can expect:
\begin{itemize}
\item \textbf{Development of Advanced Zonal PDMs:} PDMs will become more sophisticated, incorporating features like integrated voltage regulation, self-diagnostics, and communication capabilities.
\item \textbf{Software-Defined Power Management: }Software will play a bigger role in managing power distribution across zones, optimizing energy use and adapting to changing demands.
\end{itemize}
\subsection{Weight of wiring harness}
Zonal architecture contribute to reducing the complexity and weight of wiring harnesses in modern vehicles. Some advantages worth noting:
\begin{itemize}
   \item \textbf{Divided and Conquer: }The vehicle is divided into Each zone has its own dedicated processing unit and power distribution module
   \item \textbf{Shorter Runs, Less Weight:} With PDMs located within zones, cable runs become much shorter, significantly reducing the overall weight and complexity of the wiring harness.
   \item \textbf{Modular Design:} Adding new features or ECUs within a zone requires less extensive wiring modifications compared to the centralized approach, allowing for easier upgrades and customization.
\end{itemize}

\section{Future Scope}
The implementation of zonal architectures in automotive systems offers exciting opportunities for further research and development:
\begin{itemize}
\item \textbf{Edge Computing Advancements:}
The continuous development of more powerful and efficient edge computing devices specifically designed for automotive applications will be critical. This progress will enable more sophisticated processing capabilities within each zone.
\item \textbf{AI Integration within Zonal Controllers:}
Integrating more sophisticated artificial intelligence algorithms within zonal controllers could enhance real-time decision-making capabilities and improve the overall performance of autonomous driving systems.
\item \textbf{Enhanced Security Measures:}
As vehicles become more connected and reliant on complex software systems, robust cybersecurity measures for zonal architectures are becoming increasingly crucial. Future research should focus on advanced encryption techniques and intrusion detection systems specifically tailored for automotive zonal networks.
\item \textbf{Standardization Efforts:}
Industry-wide standardization of zonal architecture protocols and interfaces is essential for future progress. This would facilitate the seamless integration of components from different manufacturers and streamline the development process.

\item \textbf{Power Optimization:}
Further research on power management strategies within zonal architectures could lead to significant improvements in energy efficiency. This includes exploring advanced battery technologies and intelligent power distribution systems.

\item \textbf{Human-Machine Interface (HMI) Integration:}
Investigating how zonal architectures can improve the integration and performance of advanced HMI systems, potentially leading to more intuitive and responsive user experiences in vehicles.

\item \textbf{Adaptive Diagnostics:}
Development of more sophisticated diagnostic tools and procedures support for zonal architectures. This could include AI-driven predictive maintenance systems that can anticipate potential failures within specific zones.

\end{itemize}

\section{Conclusion}

Zonal architecture represents a transformative approach in automotive design, addressing the growing complexity of modern vehicles and offering numerous advantages over traditional centralized systems, particularly those equipped with advanced driver assistance systems (ADAS) and autonomous driving capabilities. This paper has explored the shift in automotive architecture from traditional centralized systems to distributed zonal approaches, highlighting their significant impact on scalability, reliability, vehicle performance, and cost-effectiveness. The integration of neural networks and edge computing within zonal architectures has developed as a crucial factor in enabling sophisticated sensor fusion and real-time decision-making for autonomous vehicles. These technologies work synergistically to enhance object recognition, scene interpretation, steering and navigation, and route planning optimization, ultimately leading to improved safety and efficiency in autonomous driving systems.

Key advantages of zonal architecture include: 
\begin{itemize}
\item Performance improvement through distributed processing and reduced data bottlenecks
\item Enhanced scalability and flexibility, allowing for easier integration of new features and sensors
\item Increased reliability due to redundancy and localized fault isolation
\item Cost-effectiveness resulting from simplified wiring and optimized hardware utilization.

The influence of zonal architectures extends beyond performance improvements, significantly affecting vehicle diagnostics, power distribution, and management systems. This approach enhances diagnostic processes, enables more efficient troubleshooting, and supports the implementation of over-the-air (OTA) updates. Zonal architectures allow for smarter distribution strategies, including the use of smart fuses and improved voltage regulation, leading to enhanced energy efficiency and reliability in the power management field. However, challenges remain in implementing zonal architectures effectively. These include addressing heat management concerns in distributed power systems, seamless integration of multiple ECUs, and maintaining robust cybersecurity measures. The automotive sector is actively working to overcome these problems through collaborative efforts between OEMs, chipmakers, and software developers. Zonal architectures will play an increasingly crucial role in higher levels of autonomy and connectivity. The future of automotive design lies in the continued refinement of these architectures, with a focus on advanced AI integration, standardization, and the development of efficient edge computing devices for automotive applications.

In conclusion, zonal architecture represents not just a step forward but a revolutionary leap in automotive technology. It provides the foundation necessary for the next generation of intelligent, efficient, and safer vehicles, paving the way for a future where autonomous driving becomes a reality. As research and development in this field progress, we can expect more innovative applications of zonal architectures, further transforming the automotive landscape and enhancing the driving experience for end users.
\end{itemize}

\end{document}